\pdfoutput=1

\documentclass[11pt]{article}

\usepackage[]{acl}

\usepackage{times}
\usepackage{latexsym}

\usepackage[T1]{fontenc}

\usepackage[utf8]{inputenc}

\usepackage{microtype}

\usepackage{inconsolata}

\usepackage{hyperref}
\usepackage{url}

\usepackage{hyperref}
\usepackage{url}
\usepackage{epsfig}
\usepackage{amsmath}
\usepackage{amssymb}
\usepackage{microtype}
\usepackage{xspace}
\usepackage{subfig}
\usepackage{graphicx,wrapfig}
\usepackage{booktabs}
\usepackage{makecell}
\usepackage{framed}
\usepackage{multirow}
\usepackage{enumitem}
\usepackage{bm}
\usepackage{pifont}
\newcommand{\cmark}{\ding{51}}%
\newcommand{\xmark}{\ding{55}}%

\newcommand{\para}[1]{\paragraph{\textbf{#1}}}
\newcommand{\data}{\textsc{WinoViz}\xspace}

\title{\data: Probing Visual Properties of Objects Under Different States}

\author{
    Woojeong Jin,~
    Tejas Srinivasan,~
    Jesse Thomason,~
    Xiang Ren\\
    Department of Computer Science, University of Southern California, USA \\
    {\texttt{\{woojeong.jin,tejas.srinivasan,jessetho,xiangren\}@usc.edu}}\\
}

\begin{document}

\maketitle
\begin{abstract}

Humans perceive and comprehend different visual properties of an object based on specific contexts. 
For instance, we know that a banana turns \textit{brown} ``when it becomes rotten,'' whereas it appears \textit{green} ``when it is unripe.'' 
Previous studies on probing visual commonsense knowledge have primarily focused on examining language models' understanding of \textit{typical} properties (e.g., colors and shapes) of objects. 
We present \data, a text-only evaluation dataset, consisting of 1,380 examples that probe the reasoning abilities of language models regarding variant visual properties of objects under different contexts or states. 
Our task is challenging since it requires pragmatic reasoning (finding intended meanings) and visual knowledge reasoning. 
We also present multi-hop data, a more challenging version of our data, which requires multi-step reasoning chains to solve our task. 
In our experimental analysis, our findings are: 
a) Large language models such as GPT-4 demonstrate effective performance, but when it comes to multi-hop data, their performance is significantly degraded. 
b) Large models perform well on pragmatic reasoning, but visual knowledge reasoning is a bottleneck in our task. 
c) Vision-language models outperform their language-model counterparts. 
d) A model with machine-generated images performs poorly in our task. This is due to the poor quality of the generated images.

\end{abstract}

\section{Introduction}

Language models (LMs) have long struggled with the challenge of developing intuitive reasoning abilities and acquiring knowledge from experience, despite these being innate for humans. 
Humans effortlessly enhance their knowledge by observing the visual world through their eyes. However, obtaining this type of knowledge presents difficulties because it is often not explicitly described in text form. Overcoming these challenges necessitates visual grounding, which involves establishing connections and associations between language and visual information to facilitate comprehension and interpretation of the visual world.

\begin{figure}[t!]
    \centering
    \includegraphics[width=0.98\linewidth]{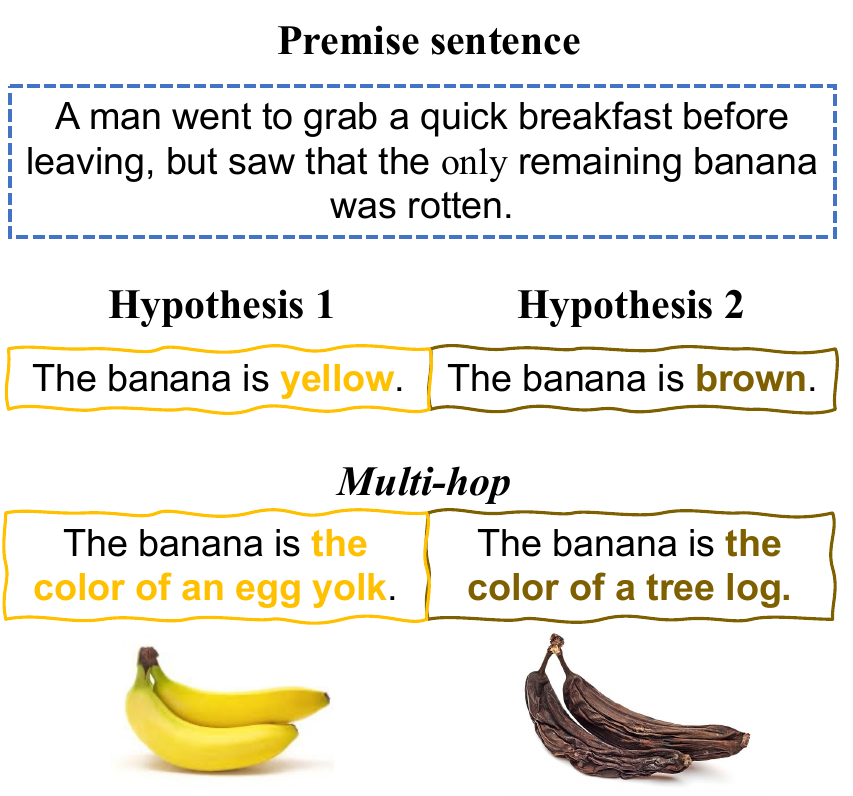}
    \caption{\textbf{The \data task.} We investigate the divergent properties of an object and explore the reasoning abilities of language models pertaining to object attributes. The premise sentence depicts a scene involving a banana and two hypothesis sentences describe the visual properties of a banana. The task is to choose a more plausible hypothesis given the premise. For the multi-hop version, we replace the visual attribute word with another object word which has a similar visual attribute.}
    \label{fig:data}
\end{figure}

Previous studies have predominantly aimed at investigating language models in relation to object prototypical visual properties such as color, shape, and affordance, and transferring such knowledge from vision-language models~\citep{norlund2021transferring,paik2021world,zhang2022visual,li2023can}. These studies discovered that reporting bias has a negative impact on model performance, but multimodal training can alleviate these effects. 
However, these studies are limited in that they mainly focused on typical properties such as color and shape.
In this work, we study language models' reasoning ability on associations between objects and their visual properties across different object states.
The task requires a model to reason about different states of an object where the object may exhibit different properties.



In this work, we investigate the divergent properties of an object and explore the reasoning abilities of language models pertaining to object attributes.
To accomplish this goal, we compose a unique, text-only evaluation dataset named \data, meticulously compiled through crowdsourcing efforts. For instance, we assigned an annotator with the task of crafting a premise sentence that portrays a scene involving a banana, along with two hypothesis sentences highlighting its visual properties, as depicted in Fig.~\ref{fig:data}. The premise sentence should demonstrate better compatibility with one of the hypothesis sentences compared to the other.
In order to select a more plausible hypothesis sentence, one must comprehend the properties of a banana under different states.
Additionally, we introduce a more challenging version of the dataset, referred to as the multi-hop data, which requires multi-step reasoning chains to solve our task.
For the multi-hop version, we replace the visual attribute word with another object word that shares a similar visual attribute.


We utilize a benchmark to assess the zero-/few-shot performance of various language models, which encompass both text-only models and vision-augmented LMs.
In our examination of text-only language models, we consider a range of models, including BERT~\cite{kenton2019bert}, T5~\cite{raffel2020exploring,chung2022scaling}, and models from the GPT family~\cite{brown2020language}. These models vary in scale, with parameters spanning from 110 million to 175 billion.
In addition to text-based models, we explore models that incorporate visual information, such as VL-BERT~\cite{su2019vl} and Oscar~\cite{li2020oscar}. Our motivation for this exploration is the notion that acquiring visual knowledge from images can enhance the capabilities of language models. 
Furthermore, we leverage machine-generated images~\cite{rombach2022high} to guide the LMs inspired by imagination-guided text generation~\cite{zhu2022visualize}.




In our experiments with the \data benchmark, we have observed the following key findings:
a) Large language models, such as GPT-4, demonstrate effective performance. However, when it comes to multi-hop data, their performance is significantly degraded.
b) Large models perform well on pragmatic reasoning, but visual knowledge reasoning is a bottleneck in our task.
c) Vision-language models outperform their language-model counterparts.
d) A model with machine-generated images performs poorly in our task. This is due to the poor quality of the generated images.



\section{The \data Task}


We present the proposed \data task using precise mathematical notations and address its inherent challenges. The \data task entails the need for a model to deduce whether objects can demonstrate prototypical behaviors in various scenarios. More precisely, when provided with a natural language sentence describing an object engaged in a particular behavior (\textit{premise sentence}), the model must determine between two sentences presenting contrasting visual attributes of the object (\textit{hypothesis sentences}).



\para{Task Definition.}
Formally, the input of the \data task is a premise sentence $p$ about an object $o$ and two hypothesis sentences $\{c_1, c_2\}$ describing contrasting attributes about the object.
The premise sentence describes a common scenario about an object in our daily life and the hypothesis sentences describe the visual characteristics of the object. 
A scenario can depict either a static situation or a short series of actions.
The expected output is one of the hypotheses that is more compatible with the premise sentence than the other.
Furthermore, we propose two different tasks: \textit{single-hop} and \textit{multi-hop} reasoning data.

\para{Challenges.}
The \data  task assesses a machine's reasoning ability regarding objects in our daily lives, focusing on the varied properties of these objects. One particular area where models frequently encounter challenges is in grasping visual knowledge related to common objects and their attributes. This difficulty arises because such knowledge is seldom explicitly detailed in the training text, primarily due to reporting bias~\cite{norlund2021transferring,jin2022leveraging}.
Furthermore, our task is challenging since it requires pragmatic reasoning and visual knowledge reasoning. 
Pragmatic reasoning involves finding intended meanings in the text, while visual knowledge reasoning requires a model to reason about the properties of objects.
We also introduce a more challenging version of \data called multi-hop data, which necessitates multi-step reasoning chains to solve our task.

\section{The \data Data}
In this section, we describe how we construct our \data dataset.

\subsection{Data Collection}
The data collection is broken down into three sections: (1) collecting candidate objects, (2) annotating premise and hypothesis sentences, (3) verifying the quality of the annotated dataset, and (4) human evaluation.

\begin{figure}[t]
    \centering
    \includegraphics[width=0.98\linewidth]{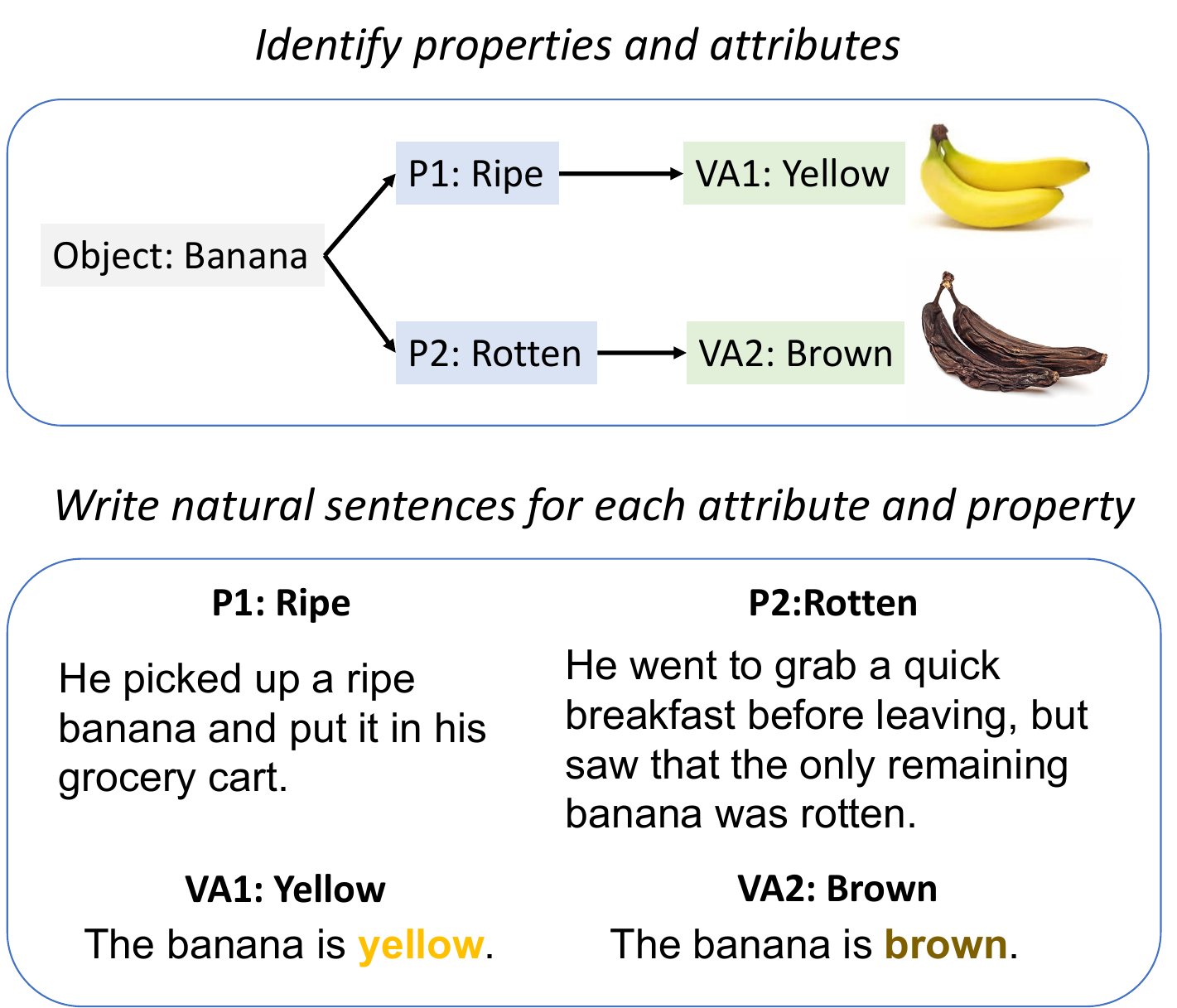}
    \caption{\textbf{Dataset Collection.} We collect our data through crowdsourcing efforts. The first step is to identify properties and visual attributes for an object and the second step is to write natural sentences for each property and attribute. Sentences with properties will be used as premise sentences and sentences with visual attributes will be used as hypothesis sentences.}
    \label{fig:datacol}
\end{figure}

\para{Object Collection.}
To begin with, we gather a collection of objects along with their potential properties or attributes for constructing our data. These objects and attributes are obtained by scraping information from reliable sources such as Memory Colors~\citep{norlund2021transferring}, Visual Property Norms~\citep{hagstrom2022models}, and McRae feature norms~\citep{mcrae2005semantic}.
Through this process, we manage to collect a total of 800 unique objects and 302 unique attributes. However, it is necessary to refine our dataset by filtering out attributes that are either too abstract or non-visual in nature. To accomplish this, we employe specific heuristics to ensure the inclusion of only concrete and visually relevant attributes.
As a result of this filtering process, we successfully obtain a final dataset comprising 775 objects and 156 attributes.

\para{Dataset Annotation.}
We utilized Amazon Mechanical Turk~\cite{crowston2012amazon} for data annotation, as depicted in Figure~\ref{fig:datacol}. The data annotation process involves several steps. Initially, annotators are given an object, and are instructed to identify two properties for the object and corresponding visual attributes for those properties. For example, for the object \textit{banana}, the annotator may come up with two properties \textit{ripe} and \textit{rotten}, which would have corresponding visual attributes \textit{yellow} and \textit{brown}, respectively. 
After identifying a pair of object properties and visual attributes, they are tasked with composing natural language sentences for each attribute and property. 
The properties are associated with premise sentences, while the attributes were linked to hypothesis sentences.

Annotators were selected from a small pool of Mechanical Turkers that the authors had previously worked with. 
The Turkers had to further pass a qualification task that tested their understanding of the annotation task. 
The authors manually examined the annotations to ensure quality of the collected data.



\subsection{Versions of \data}
We now collect our \data data. 
We also propose the multi-hop data, a more challenging version of \data, and a dataset for probing visual knowledge.
For the multi-hop data, we create new hypothesis options that require more intermediate steps while we simplify the premise sentences to measure the ability of models about visual knowledge.


\para{Multi-hop Data.}
To create a more challenging task, we introduce a multi-hop version of our data, which requires more intermediate steps.
The basic idea of the multi-hop data is to replace a visual attribute word in hypotheses with another object word which has a similar visual attribute. This requires one more reasoning step to find out the visual attribute.
For example, one hypothesis option is `The banana is yellow.'. Then 'yellow' can be replaced with `the color of an egg yolk.'
So the new hypothesis option for the multi-hop version is 'The banana is the color of an egg yolk.'
The multi-hop version is more challenging since a model has to find out what color is an egg yolk.
We focus on color, shape, material on the multi-hop data and curate prototypical objects for each visual property word.
We get 200 samples for the multi-hop data.

\para{Pragmatic Reasoning vs. Visual Knowledge Reasoning.}

Another important aspect of this work is that models genuinely understand and know visual knowledge. 
Our task requires pragmatic reasoning, the process of finding the intended meaning, and visual knowledge reasoning but models may fail in one of the reasoning steps. 
Thus, we decouple the premise sentence into pragmatic reasoning step and visual knowledge reasoning step to analyze which step is a bottleneck.
Pragmatic reasoning involves finding the intended meaning and finding key phrases for the next step, visual knowledge reasoning.
For example, a model should first find `the banana is ripe' given the premise sentence in the pragmatic reasoning step (Figure~\ref{fig:data}). Given the simplified sentence, a model should choose a better option, `the banana is yellow', in the visual knowledge reasoning step.
We obtain 160 samples to study this (Section~\ref{sec:exp:prag}).


\section{Experiments}

We first describe the experimental setup used in our analysis and share experimental results.

\begin{table}[tb!]
\centering
\resizebox{0.98\linewidth}{!}{
\begin{tabular}[t]{l|c|c|c}
\toprule
\textbf{Model}   & \textbf{\# Params} &\textbf{Public}    &\textbf{VL model}      \\
\midrule
BERT-Base   & 109M  & \cmark   & \xmark  \\
BERT-Large  & 335M  & \cmark   & \xmark  \\
VL-BERT-Large   & 335M  & \cmark   & \cmark  \\
Oscar-Large & 335M  & \cmark   & \cmark  \\
CLIP-Large  & 427M  & \cmark   & \cmark  \\
FLAN-T5-XXL & 11B   & \cmark   & \xmark  \\
InstructBLIP    & 11B   & \cmark   & \cmark  \\
LLaMA2  & 13B   & \cmark   & \xmark  \\
LLaVA   & 13B   & \cmark   & \cmark  \\
GPT-3    & 175B  & \xmark   & \xmark  \\
GPT-3.5  & Unknown   & \xmark   & \xmark  \\
GPT-4    & Unknown   & \xmark   & \xmark  \\
\bottomrule
\end{tabular}}
\caption{\textbf{A list of models used in the experiments:} BERT~\cite{kenton2019bert}, CLIP~\cite{radford2021learning}, VL-BERT~\cite{su2019vl}, Oscar~\cite{li2020oscar}, FLAN-T5~\cite{chung2022scaling}, InstructBLIP~\cite{DBLP:journals/corr/abs-2305-06500}, LLaMA2~\cite{touvron2023llama}, LLaVA~\cite{liu2023visual}, GPT-3~\cite{brown2020language,ouyang2022training}, and GPT-4~\cite{DBLP:journals/corr/abs-2303-08774}. We use the `text-davinci-003' API for GPT-3, `gpt-3.5-turbo-instruct' for GPT-3.5, and `gpt-4-0314' for GPT-4.}
\label{tab:models}
\end{table}

\paragraph{\textbf{Language Models.}}
We experiment with 7 language models in total (Table~\ref{tab:models}). We include encoder-only, encoder-decoder, decoder-only models. The sizes of LMs vary from 109M to 175B.
We include large LMs, GPT-3, GPT-3.5, and GPT-4~\cite{brown2020language,ouyang2022training,DBLP:journals/corr/abs-2303-08774}.

\paragraph{\textbf{Vision-language Models.}}
We also experiment with 5 vision-language models in total (Table~\ref{tab:models}).
Our task requires visual knowledge on objects under different states. 
Such knowledge can be obtained from image-caption datasets and thus we explore vision-language models.
We analyze whether vision-language models can outperforms language models on our task.
For model evaluation, we deliberately exclude any image inputs and refrain from utilizing the image encoders of the models. Instead, we focus solely on the language components of the models. 
We use encoder-only models, VL-BERT~\cite{su2019vl} and Oscar~\cite{li2020oscar}, and a decoder-only model, LLaVA-v1.5~\cite{liu2023visual}, and a bi-encoder model, CLIP (`clip-vit-large-patch14')~\cite{radford2021learning}.
Additionally, we utilize an image generation approach, Stable Diffusion~\cite{rombach2022high}, to generate images.
We use the generated images to guide the LMs inspired by imagination-guided text generation~\cite{zhu2022visualize}.

\paragraph{\textbf{Inference.}}
In our analysis, we rely on zero-shot inference and few-shot in-context learning for encoder-decoder, decoder-only models. 
Our prompt design for the zero-shot inference is as follows:
\textit{``You will be given a sentence, and two options. Output either Option 1 or Option 2, depending on which option is more likely to be true given the sentence.''}
For the few-shot in-context learning, we use 4 examples.
We also adopt chain-of-thought prompting~\cite{wei2022chain} for the few-shot inference.
In addition to the encoder-decoder and decoder-only models, we explore encoder-only models.
Encoder-only models cannot do zero-shot inference for multi-choice tasks since it requires a task-specific head for unseen tasks. 
Thus, we fine-tune the encoder-only models with SNLI~\cite{bowman2015large} and ANLI~\cite{nie2019adversarial} datasets and we use only `contradiction' and `entailment' labels in fine-tuning.

\paragraph{\textbf{Evaluation Setup.}}
We evaluate models with two different metrics: individual accuracy (Ind.) and pair accuracy (Pair).
Individual accuracy refers to accuracy on each individual question, while pair accuracy refers to the accuracy on each pair of questions.
In \data, two premise sentences are paired and they share the same set of hypothesis options.
We measure the model's performance based on its ability to accurately predict both premise sentences. If the model's prediction is correct for only one of the premise sentences in the pair, we consider the prediction less robust.


\subsection{Analysis Questions}
In our empirical analysis, we try to answer the following questions:
\begin{enumerate}
    \item How good are large models on our task? When it comes to multi-hop data, how good are they? (Section~\ref{sec:exp:zero})
    \item Do few-shot prompting and CoT prompting improve the results? (Section~\ref{sec:exp:few})
    \item Which reasoning step between pragmatic reasoning and visual knowledge reasoning is main bottleneck in our task? (Section~\ref{sec:exp:prag})
    \item Do vision-language models outperform language-model counterparts? (Section~\ref{sec:exp:zero}) 
    \item Can we improve performance using image generation approaches? Do generated images help solving our task? (Section~\ref{sec:exp:genimg})
\end{enumerate}


\begin{table}[tb!]
\centering
\resizebox{\linewidth}{!}{
\begin{tabular}[t]{l|cccc}
\toprule
    \multirow{2}{*}{\textbf{Model} }   & \multicolumn{2}{c}{\textbf{Single-hop}} &\multicolumn{2}{c}{\textbf{Multi-hop}}     \\
    \cmidrule(lr){2-3} \cmidrule(lr){4-5}
    & Ind. & Pair & Ind. & Pair  \\
\midrule
FLAN-T5-XXL & 86.24   & 72.71   & 68.09   & 40.43     \\
LLaMA2      & 73.28   & 48.85   & 52.84   & 20.45     \\
LLaVA       & 79.47   & 59.63   & 56.82   & 17.05     \\
GPT-3       & 84.17   & 69.24   & 58.5   & 22     \\
GPT-3.5     & 86.58   & 75.62   & 58   & 20     \\
GPT-4       & 90.25   & 81.19   & 72   & 45     \\
\bottomrule
\end{tabular}}
\caption{\textbf{Results on \data in a zero-shot manner.} We evaluate large models  using 0 examples on both our single-hop and multi-hop datasets. We observe that these models performed well on the single-hop data; however, their performance is significantly degraded on the multi-hop data.
}
\label{tab:result}
\end{table}

\subsection{Zero-shot Results}
\label{sec:exp:zero}
We evaluate language models and vision-language models in a zero-shot way, without utilizing any training data (Table~\ref{tab:result}).
Overall, large models perform well on the single-hop data, but their performance is significantly degraded on the multi-hop data.
Among them, GPT-4 exhibits the best overall performance on both single-hop and multi-hop tasks.
Surprisingly, FLAN-T5-XXL, the smallest model among the comparison, yields comparable results to larger models, including GPT-3. Moreover, it outperforms GPT-3 and GPT-3.5 on the multi-hop dataset.
LLaVA, built upon LLaMA2 and trained with image-caption datasets, shows noteworthy performance. As indicated in the table, LLaVA surpasses LLaMA2 on both single-hop and multi-hop data, suggesting that image-caption datasets enhance reasoning in our task.

\begin{table}[tb!]
\centering
\resizebox{\linewidth}{!}{
\begin{tabular}[t]{l|cccc}
\toprule
    \multirow{2}{*}{\textbf{Model} }   & \multicolumn{2}{c}{\textbf{Single-hop}} &\multicolumn{2}{c}{\textbf{Multi-hop}}     \\
    \cmidrule(lr){2-3} \cmidrule(lr){4-5}
    & Ind. & Pair & Ind. & Pair  \\
\midrule
FLAN-T5 (0)     & 86.35   & 73.17   & 68.09   & 40.43     \\
FLAN-T5 (4)     & 87.84   & 76.15   & 69.32   & 42.05     \\
FLAN-T5 (4 CoT) & 87.16   & 74.77   & 67.05   & 38.64     \\
GPT-3.5 (0)         & 86.58   & 75.62   & 58   & 20     \\
GPT-3.5 (4)         & 88.42   & 77.75   & 62.5   & 28.41     \\
GPT-3.5 (4 CoT)     & 77.18   & 59.63   & 65.34   & 34.09     \\
\bottomrule
\end{tabular}}
\caption{\textbf{Results on \data with 4-shot in-context learning.} We use FLAN-T5-XXL and GPT-3.5 in this analysis. Standard prompting marginally improves the performance of them, while chain-of-thought prompting is beneficial for GPT-3.5 in the multi-hop task.
}
\label{tab:result_few}
\end{table}

\subsection{Few-shot Results}
\label{sec:exp:few}
Table~\ref{tab:result_few} displays the results with 4 in-context examples for FLAN-T5-XXL and GPT-3.5. 
We conduct tests using standard prompting and chain-of-thought prompting in this experiment. 
Initially, standard prompting with 4 in-context examples marginally improves the performance of FLAN-T5 and GPT-3.5 on both single-hop and multi-hop tasks. It's surprising that chain-of-thought prompting appears to negatively impact the performance of GPT-3.5. However, it proves beneficial for GPT-3.5. in the multi-hop task. We speculate that the effectiveness of chain-of-thought prompting increases when the task is more challenging.

\begin{table}[tb!]
\centering
\resizebox{\linewidth}{!}{
\begin{tabular}[t]{l|cccc}
\toprule
    \multirow{2}{*}{\textbf{Method} }   & \multicolumn{2}{c}{\textbf{Single-hop}} &\multicolumn{2}{c}{\textbf{Multi-hop}}     \\
    \cmidrule(lr){2-3} \cmidrule(lr){4-5}
    & Ind. & Pair & Ind. & Pair \\
\midrule
BERT-Large      & 67.31   & 39.44   & 54   & 16     \\
VL-BERT-Large   & 69.61   & 42.88   & 56   & 18     \\
Oscar-Large     & 72.93   & 50.22   & 64.5   & 32     \\
\bottomrule
\end{tabular}}
\caption{\textbf{Results on \data after NLI training.} We train encoder-only models on NLI datasets and choose an option by the highest probability of the `entailment' class.
}
\label{tab:result_nli}
\end{table}

\subsection{Results of Encoder-only Models}
\label{sec:exp:enc}
Encoder-only models cannot be applied to our task without fine-tuning.
Thus, we fine-tune the encoder-only models on natural language inference datasets instead.
By doing this, our task is framed into the NLI setup and choose an option by the highest probability of the `entailment' class.
We fine-tune the encoder-only models with SNLI~\cite{bowman2015large} and ANLI~\cite{nie2019adversarial} datasets and we use only `contradiction' and `entailment' labels.
Table~\ref{tab:result_nli} shows the results of encoder-only models.
VL-BERT and Oscar are BERT-based vision-language models, and they are trained on image-caption datasets.
In our experiments, we observe that the vision-language models consistently surpass the BERT model on our dataset.


\begin{table}[tb!]
\centering
\resizebox{\linewidth}{!}{
\begin{tabular}[t]{l|ccc}
\toprule
\textbf{Model}  & \textbf{Pragmatic}    & \textbf{Visual}   & \textbf{Combined} \\
\midrule
FLAN-T5-XXL & 93.04    & 82.91    & 79.75         \\
LLaMA2      & 86.71    & 70.25    & 69.62         \\
LLaVA       & 92.41    & 74.05    & 73.25         \\
GPT-3.5     & 91.14    & 82.28    & 79.75        \\
GPT-4       & 95.57    & 88.61    & 85.44       \\
\bottomrule
\end{tabular}}
\caption{\textbf{Results on pragmatic reasoning, visual knowledge reasoning, and our original data (combined).} We study different types of reasoning in our data. We report individual accuracy.
}
\label{tab:probing}
\end{table}

\subsection{Pragmatic and Visual Knowledge Reasoning}
\label{sec:exp:prag}
We investigate whether models genuinely understand visual knowledge for our task.
Our task requires pragmatic reasoning and visual knowledge reasoning.
We decouple our task into pragmatic reasoning and visual knowledge reasoning and analyze which step is a bottleneck.
Table~\ref{tab:probing} shows the results on pragmatic reasoning (pragmatic), visual knowledge reasoning (visual), and our original data (combined), utilizing the same subset.
Firstly, results on pragmatic reasoning are better than others, suggesting that large models do well on pragmatic reasoning. For example, GPT-4 achieves 95.57\% on  pragmatic reasoning.
Main bottleneck in our task is on visual knowledge reasoning;
results on visual knowledge reasoning are lower than those on pragmatic reasoning.
When comparing LLaMA2 and LLaVA, LLaVA demonstrates superior abilities in both pragmatic reasoning and visual knowledge reasoning. Interestingly, FLAN-T5-XXL performs comparably to a proprietary model, GPT-3.5, in terms of pragmatic reasoning and visual reasoning.


\begin{table}[tb!]
\centering
\resizebox{\linewidth}{!}{
\begin{tabular}[t]{l|cc}
\toprule
\textbf{Model}    & Ind. & Pair    \\
\midrule
FLAN-T5-Base (No imgs)   & 67.89 & 40.37     \\
CLIP-Large      & 64.67   & 36.46      \\
\midrule
FLAN-T5-XXL (No imgs)   & 86.24   & 72.71      \\
FLAN-T5-XXL (Captions)   & 85.83   & 71.88      \\
\midrule
InstructBLIP     & 53.21   & 22.93      \\
\bottomrule
\end{tabular}}
\caption{\textbf{Results on \data with generated images.} We use Stable Diffusion~\cite{rombach2022high} to generate 5 images per premise sentence. We adopt majority voting at inference time to choose an option. FLAN-T5-Base (No imgs) refers to a model without any generated images, with a size comparable to CLIP-Large. FLAN-T5-XXL (No imgs) refers to a model without any generated images, while FLAN-T5-XXL (Captions) refers to a model with captions generated by BLIP2 on the generated images. Instead of directly inputting images into FLAN-T5, we extract captions from the generated images and use them as additional context.  InstructBLIP uses generated images.
}
\label{tab:result_imagegen}
\end{table}

\subsection{Using Image Generation for \data Task.}
\label{sec:exp:genimg}
Another approach for our task is to utilize image generation. We generate images based on premise sentences and employ these generated images for our task. The generated images may contain useful information that assists in identifying a correct hypothesis.
Given the generated images, there are three ways to use them.
The first method involves using CLIP~\cite{radford2021learning} on both the images and hypothesis sentences to identify a superior hypothesis option. Specifically, we calculate the cosine similarity between the embedding of a generated image and the embedding of a hypothesis option, selecting the hypothesis with a higher cosine similarity score.
The second approach is to generate captions for the generated images using a caption model. Since language models cannot directly process images, we generate captions and utilize them as additional context for the task. BLIP2~\cite{li2023blip} is employed for caption generation.
The third strategy is to reframe our task as a visual question-answering task and employ a vision-language model to identify a better option. In this setup, we use InstructBLIP~\cite{DBLP:journals/corr/abs-2305-06500}.
For image generation, we use Stable Diffusion~\cite{rombach2022high}, generating 5 images per premise sentence. A better hypothesis option is determined through majority voting.

\begin{figure}[t]
    \centering
    \includegraphics[width=0.98\linewidth]{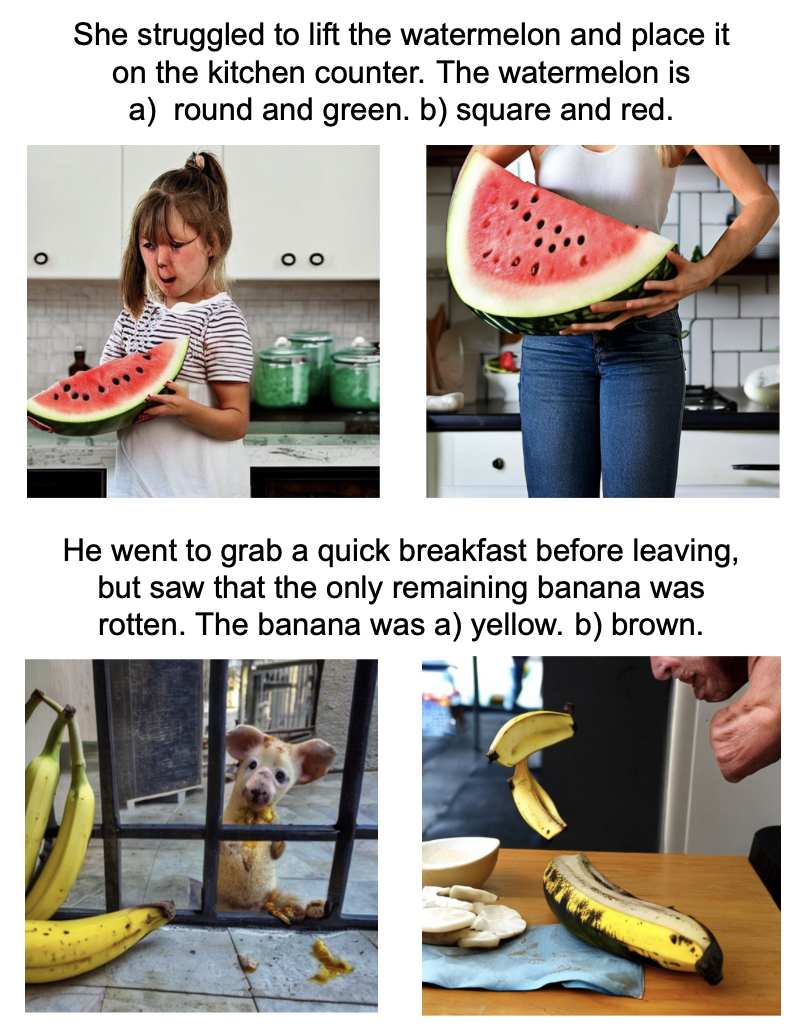}
    \caption{\textbf{Examples of generated images.} We generate images using Stable Diffusion~\cite{rombach2022high}. In the second example, the bananas in both images are yellow, leading the model to select the incorrect option. The generated image examples don't assist in selecting a more plausible hypothesis option.}
    \label{fig:genimg}
\end{figure}

Table~\ref{tab:result_imagegen} displays the outcomes related to image generation. The first approach utilizing CLIP falls short compared to FLAN-T5-Base which is slightly smaller than CLIP-Large.
In the second approach involving BLIP2 captions, we opt for FLAN-T5-XXL as the benchmark, comparing one scenario with no additional data and another incorporating captions from generated images. Our experiment reveals a notable decline in performance when captions are employed.
The third approach significantly underperforms FLAN-T5-XXL by a large margin.
These experiments collectively indicate that generated images offer limited utility for our task. Furthermore, a manual assessment of 100 generated images reveals that 66\% of them do not contribute meaningfully to our objectives.
Examples of generated images with premise sentences are shown in Figure~\ref{fig:genimg}. In the figure, the bananas in both images are yellow; the generated images do not provide any clues to choose a more plausible option.

\section{Related Work}
There are multiple perspectives on how our contributions relate to previous work, and we elaborate on this in the subsequent sections.

\paragraph{Visual Knowledge Probing.}
Several attempts have been made to assess the reasoning ability of language models regarding objects, primarily through natural language benchmarks~\citep{norlund2021transferring,hagstrom2022models,paik2021world,zhang2022visual,singh2022viphy,qasemi2021paco}. \citet{norlund2021transferring} introduced a task involving querying a multimodal model for visual commonsense knowledge related to memory colors, which are the typical colors associated with well-known objects. 
\citet{hagstrom2022models} expanded on this work by proposing visual property norms as a measure of visual commonsense knowledge in both language models and multimodal models.
\citet{paik2021world} evaluated the color perception of language models using a color dataset called CoDa, revealing that reporting bias negatively affects model performance and that multimodal training can alleviate these effects.
\citet{zhang2022visual} confirmed these findings and extended the evaluation to a wider range of visually salient properties. 
Similarly, \citet{singh2022viphy} evaluated vision-language models on a visually accessible commonsense knowledge dataset.
\citet{liu2022things} explored spatial commonsense, the knowledge about spatial position and relationship between objects, finding that image synthesis models are more capable of learning accurate and consistent spatial knowledge than other models.
\citet{gu2022language} proposed a probing dataset for physical knowledge about everyday things.
In contrast, we present a challenging dataset that probes the reasoning abilities of language models regarding variant visual properties of objects under different context.

\paragraph{Vision-Language Modeling}
Recent advances in vision-language (VL) models have led to success on vision-language tasks such as visual question answering, captioning, and grounding~\cite{antol2015vqa,lin2014microsoft,mao2016generation}.
Existing VL models jointly learn image and text representations through cross-modal alignments including VL-BERT~\cite{su2019vl}, LXMERT~\cite{tan2019lxmert}, Oscar~\cite{li2020oscar}.
Recent approaches have introduced visual instruction tuning, which involves fine-tuning a VL model using instruction-following data~\cite{liu2023visual}.

While these VL models have shown significant improvement in VL tasks, the exploration of how to transfer visual knowledge from VL modeling to language tasks remains underexplored. 
Vokenization~\cite{tan2020vokenization} utilized token-level text-to-image retrieval to transfer visual knowledge to language models. 
VidLanKD~\cite{tang2021vidlankd} employd contrastive learning to train a teacher model on video datasets and uses distillation approaches to transfer visual knowledge from the teacher to a student model.
CMKT~\cite{jin2022leveraging} investigated two types of knowledge transfer: text knowledge transfer (e.g., captions) and visual knowledge transfer (e.g., images and captions). Their findings demonstrate that such transfer can enhance performance on commonsense reasoning tasks.




\section{Conclusion}

Examining real-world object properties requires a visual understanding that language models lack. In our study, we introduced a text-only \data focused on question-answering tasks, comprising 1,380 examples exploring language models' reasoning capabilities across various visual properties of objects in diverse contexts. Our findings revealed that large language models demonstrate effective performance overall but struggle particularly with the multi-hop version of our dataset. It became apparent that the bottleneck in our task lies in the reasoning of visual knowledge. Vision-language models surpass their language-only counterparts, although image-generation approaches prove ineffective for our specific task. Future endeavors will delve into how to efficiently transfer visual knowledge from images or captions.

\section{Limitations}
Our work is focused on a specific subset of language models and vision-language models. 
We adopt vision-language models in which the language backbones are pre-trained using image-caption datasets.
Additionally, we employ Stable Diffusion for image generation, although the current output may not directly benefit our task. Utilizing state-of-the-art diffusion models could enhance image quality, yet the challenge of generating images useful for our task persists.
Moreover, our observations indicate that large language models excel in our single-hop task, achieving up to 90\% accuracy. This suggests that these large models can effectively reason over visual knowledge even in the absence of explicit visual signals. Nonetheless, how visual signals can be harnessed to enhance language models is underexplored, and we defer it to future research endeavors.




\bibliography{bib}

\appendix
\section{Appendix}



\subsection{Annotation Interfaces}
\begin{figure*}[t]
    \centering
    \includegraphics[width=0.98\linewidth]{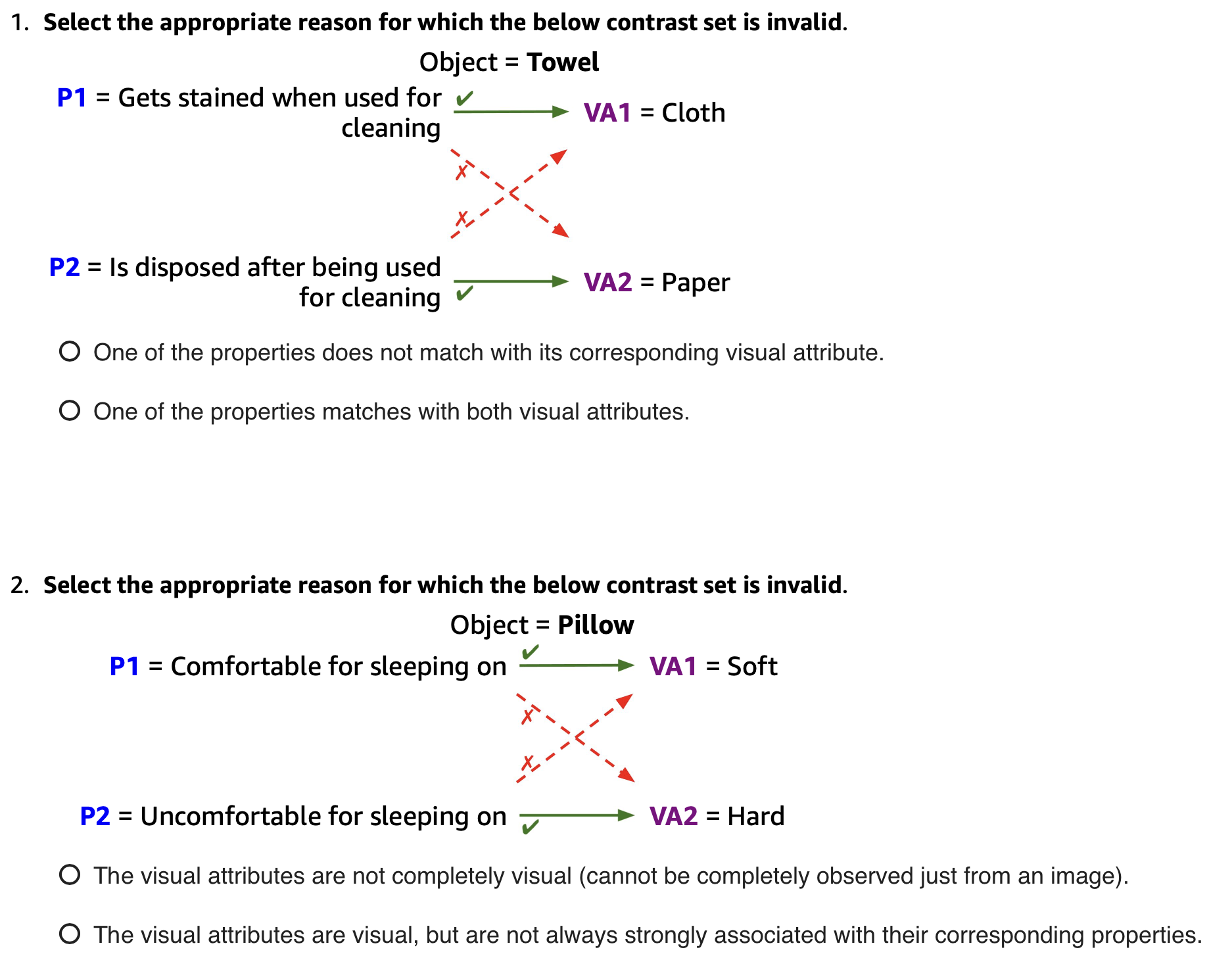}
    \caption{\textbf{The Interface of the qualification task.} We provide 12 questions to find quality workers.}
    \label{fig:qual}
\end{figure*}

\begin{figure*}[t]
    \centering
    \includegraphics[width=0.98\linewidth]{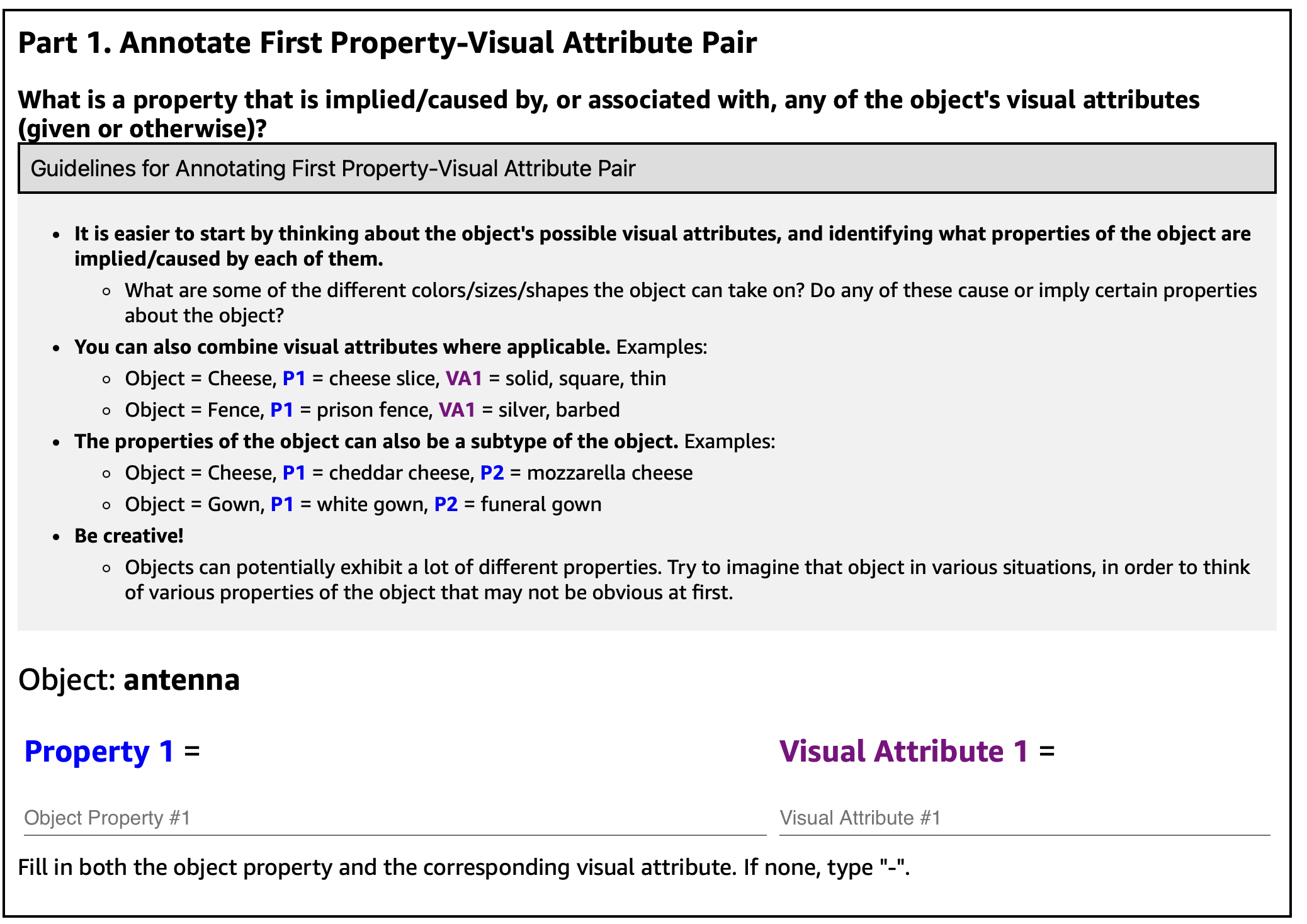}
    \includegraphics[width=0.98\linewidth]{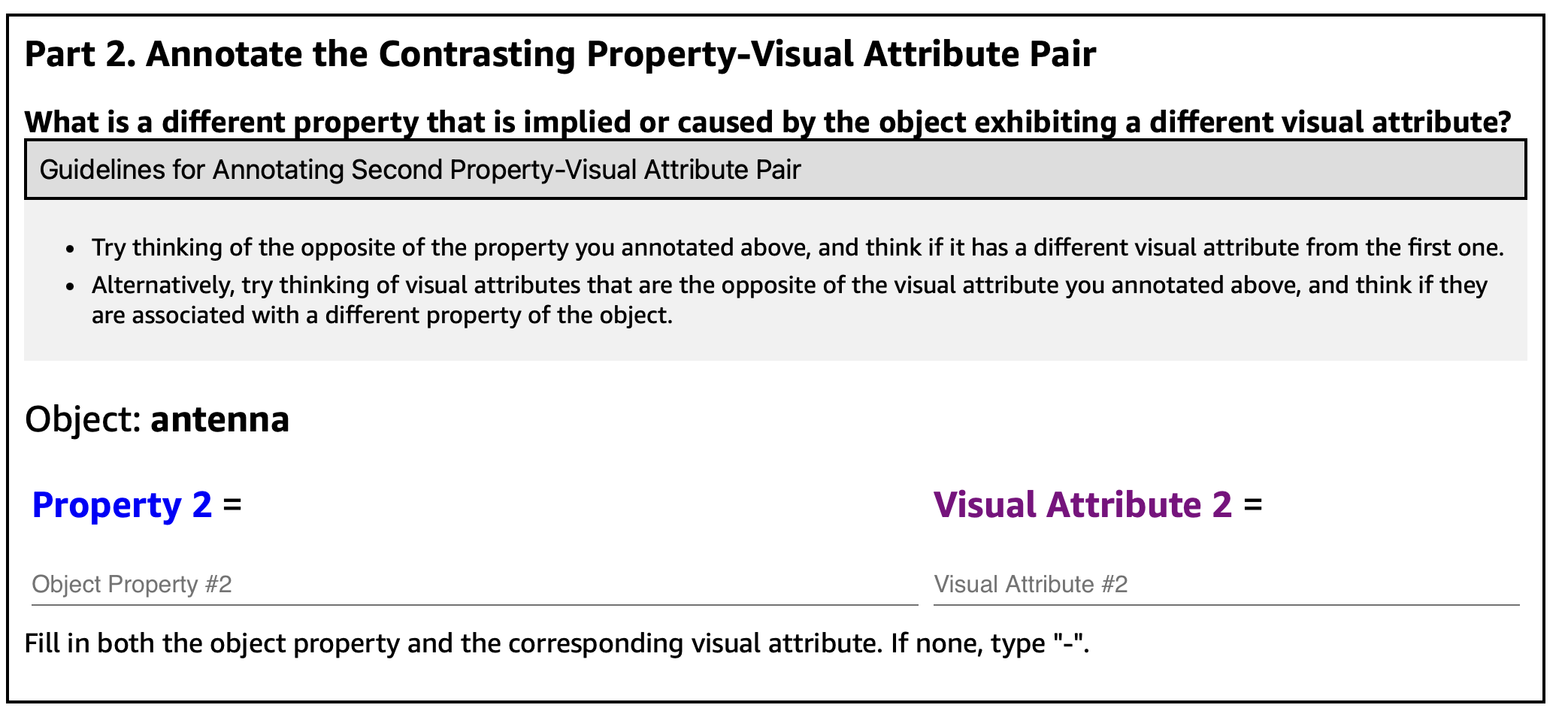}
    \caption{\textbf{Interfaces of annotating visual contrast sets (parts 1 and 2).}}
    \label{fig:task11}
\end{figure*}

\begin{figure*}[t]
    \centering
    \includegraphics[width=0.98\linewidth]{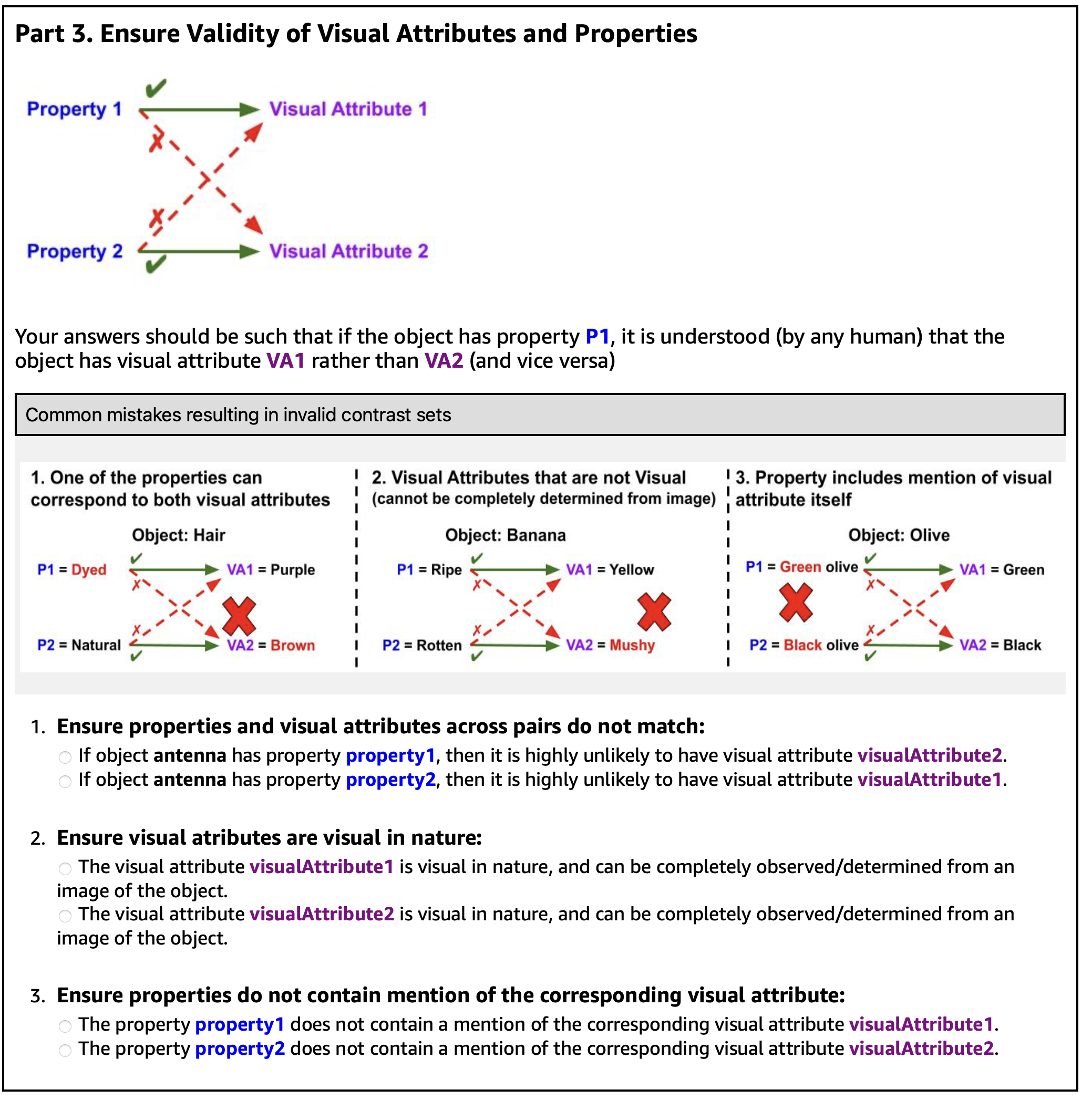}
    \caption{\textbf{Interfaces of annotating visual contrast sets (part 3).}}
    \label{fig:task12}
\end{figure*}

\begin{figure*}[t]
    \centering
    \includegraphics[width=0.98\linewidth]{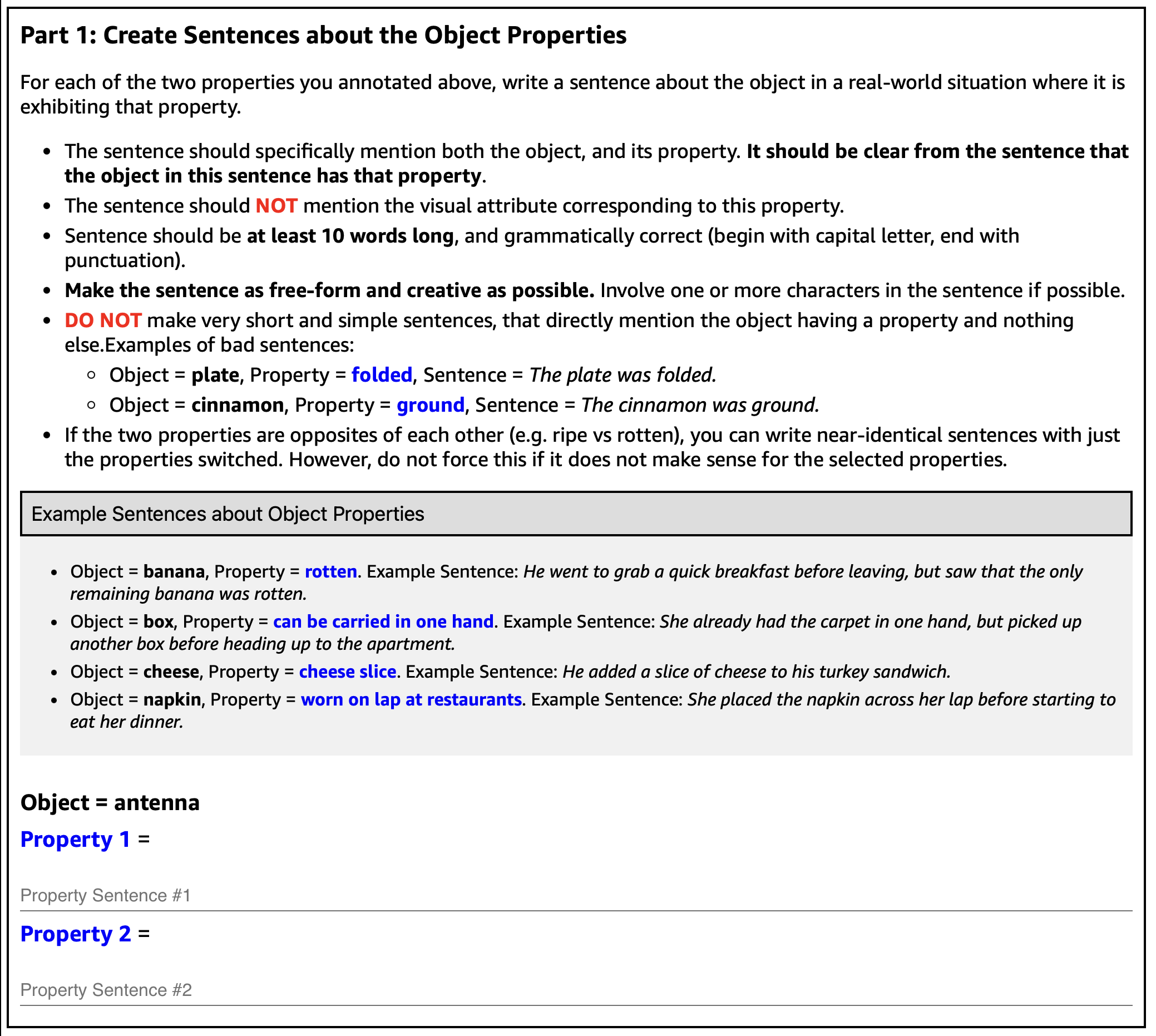}
    \includegraphics[width=0.98\linewidth]{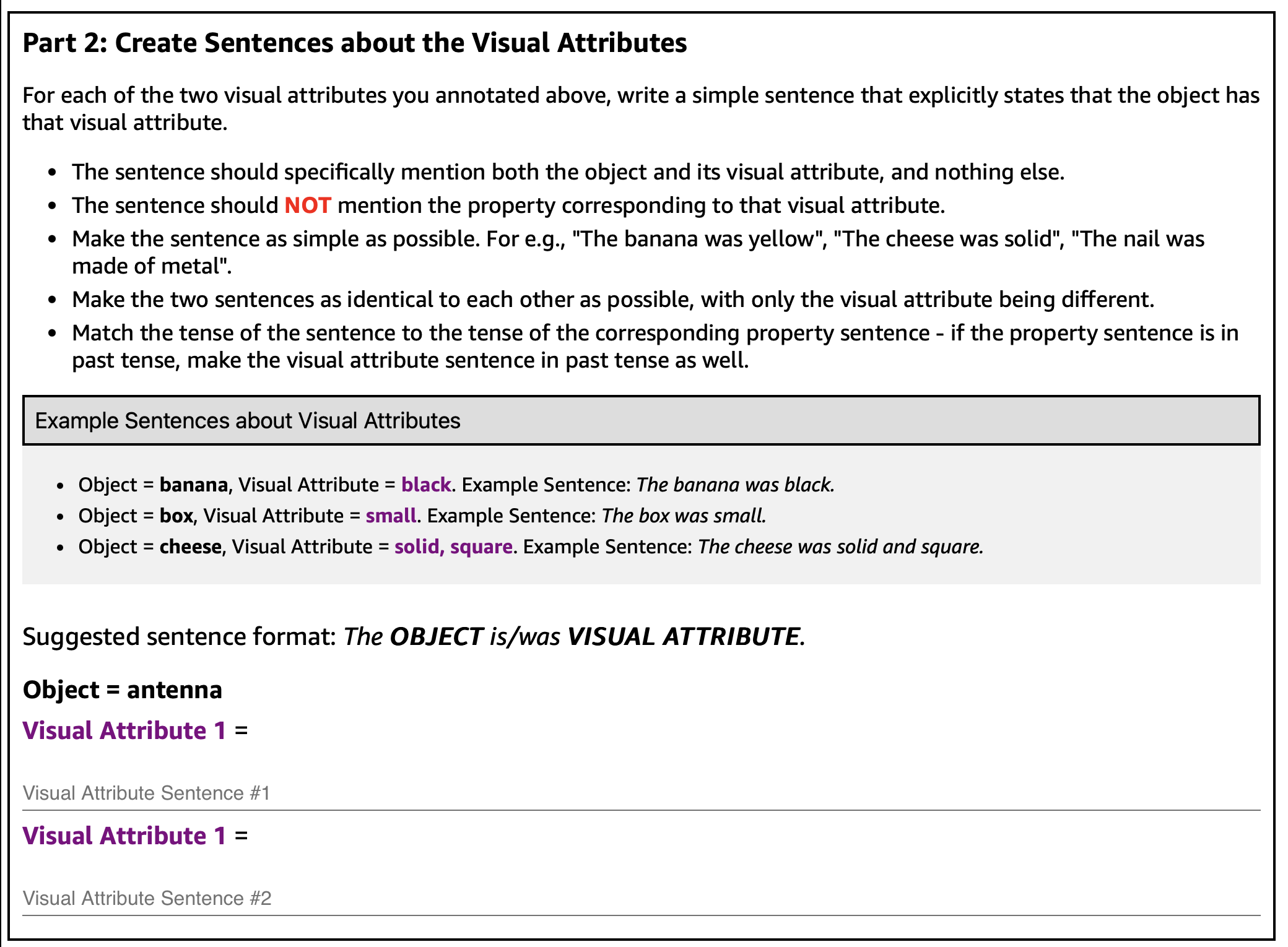}
    \caption{\textbf{Interfaces of converting contrast sets into sentence puzzles (parts 1 and 2).}}
    \label{fig:task21}
\end{figure*}

\begin{figure*}[t]
    \centering
    \includegraphics[width=0.98\linewidth]{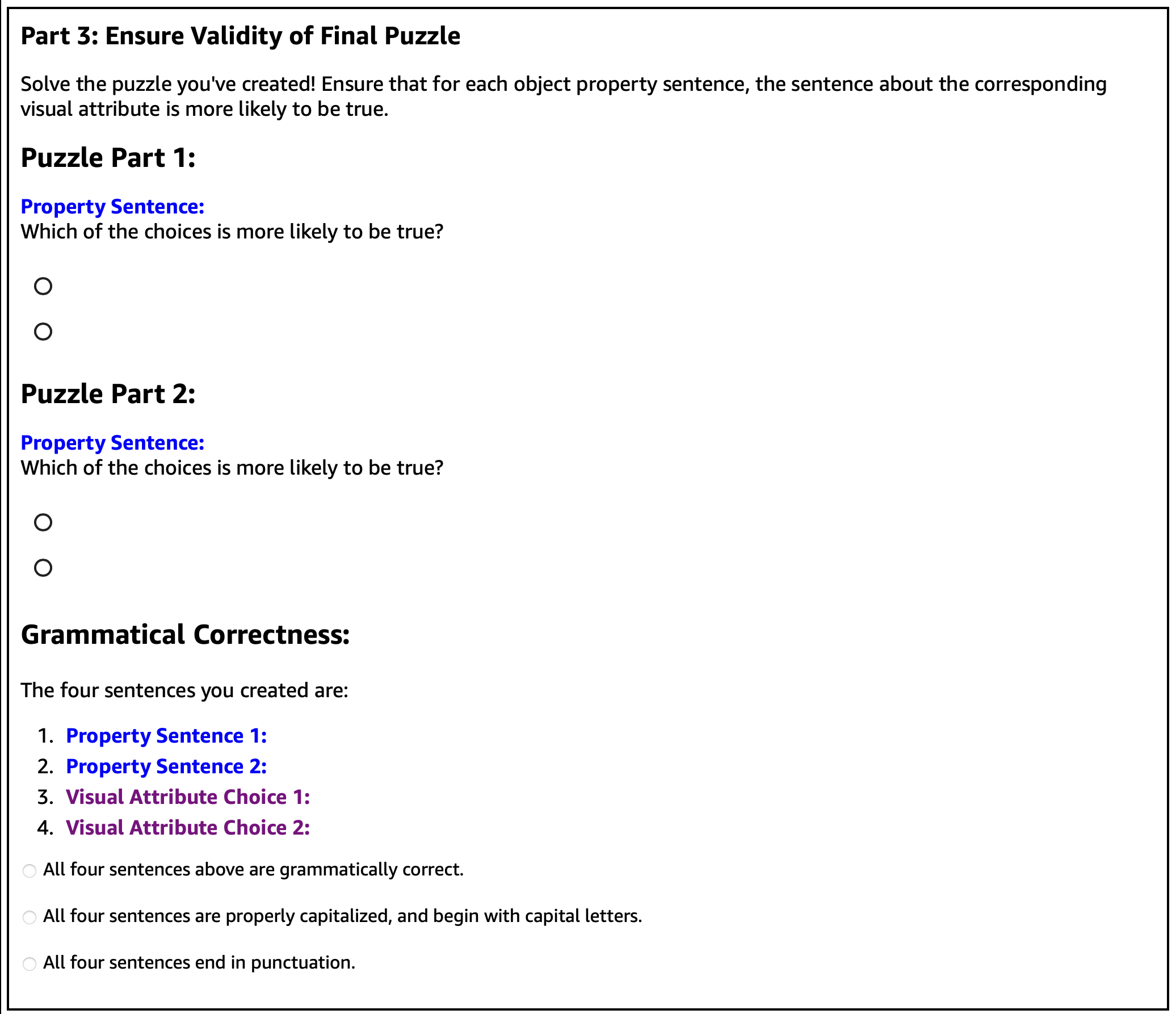}
    \caption{\textbf{Interfaces of converting contrast sets into sentence puzzles (part 3).}}
    \label{fig:task22}
\end{figure*}

We provide Turking interfaces: qualification task in Figure~\ref{fig:qual}, and annotation task in Figures~\ref{fig:task11}, \ref{fig:task12}, \ref{fig:task21}, \ref{fig:task22}.



\end{document}